\documentclass{article}

\usepackage{microtype}
\usepackage{graphicx}
\usepackage{subfigure}
\usepackage{booktabs} 
\usepackage{amsmath}

\usepackage{tabularx}
\usepackage{multirow}
\usepackage{amssymb}
\usepackage{color}
\usepackage{subfigure}

\usepackage{hyperref}
\usepackage[accepted]{icml2018}

\icmltitlerunning{Understanding the Disharmony between Dropout and Batch Normalization by Variance Shift}

\begin{document}

\twocolumn[
\icmltitle{Understanding the Disharmony between Dropout and Batch Normalization by Variance Shift}



\icmlsetsymbol{equal}{*}

\begin{icmlauthorlist}
\icmlauthor{Xiang Li}{njust}
\icmlauthor{Shuo Chen}{njust}
\icmlauthor{Xiaolin Hu}{tsinghua}
\icmlauthor{Jian Yang}{njust}
\end{icmlauthorlist}

\icmlaffiliation{njust}{DeepInsight@PCALab, Nanjing University of Science and Technology, China}
\icmlaffiliation{tsinghua}{Tsinghua National Laboratory for Information Science and Technology (TNList) Department of Computer Science and Technology, Tsinghua University, China}

\icmlcorrespondingauthor{Xiang Li}{xiang.li.implus@njust.edu.cn}

\icmlkeywords{Machine Learning, ICML}

\vskip 0.3in
]



\printAffiliationsAndNotice{}  


\begin{abstract}
	This paper first answers the question ``why do the two most powerful techniques Dropout and Batch Normalization (BN) often lead to a worse performance when they are combined together?'' in both theoretical and statistical aspects. Theoretically, we find that Dropout would shift the variance of a specific neural unit when we transfer the state of that network from train to test. However, BN would maintain its statistical variance, which is accumulated from the entire learning procedure, in the test phase. The inconsistency of that variance (we name this scheme as ``variance shift'') causes the unstable numerical behavior in inference that leads to more erroneous predictions finally, when applying Dropout before BN. Thorough experiments on DenseNet, ResNet, ResNeXt and Wide ResNet confirm our findings. According to the uncovered mechanism, we next explore several strategies that modifies Dropout and try to overcome the limitations of their combination by avoiding the variance shift risks.
\end{abstract}

\begin{figure}[t]
	\vspace{-10pt}
	\centering
	\setlength{\fboxrule}{0pt}
	\subfigure{
		\includegraphics[width=3.24in]{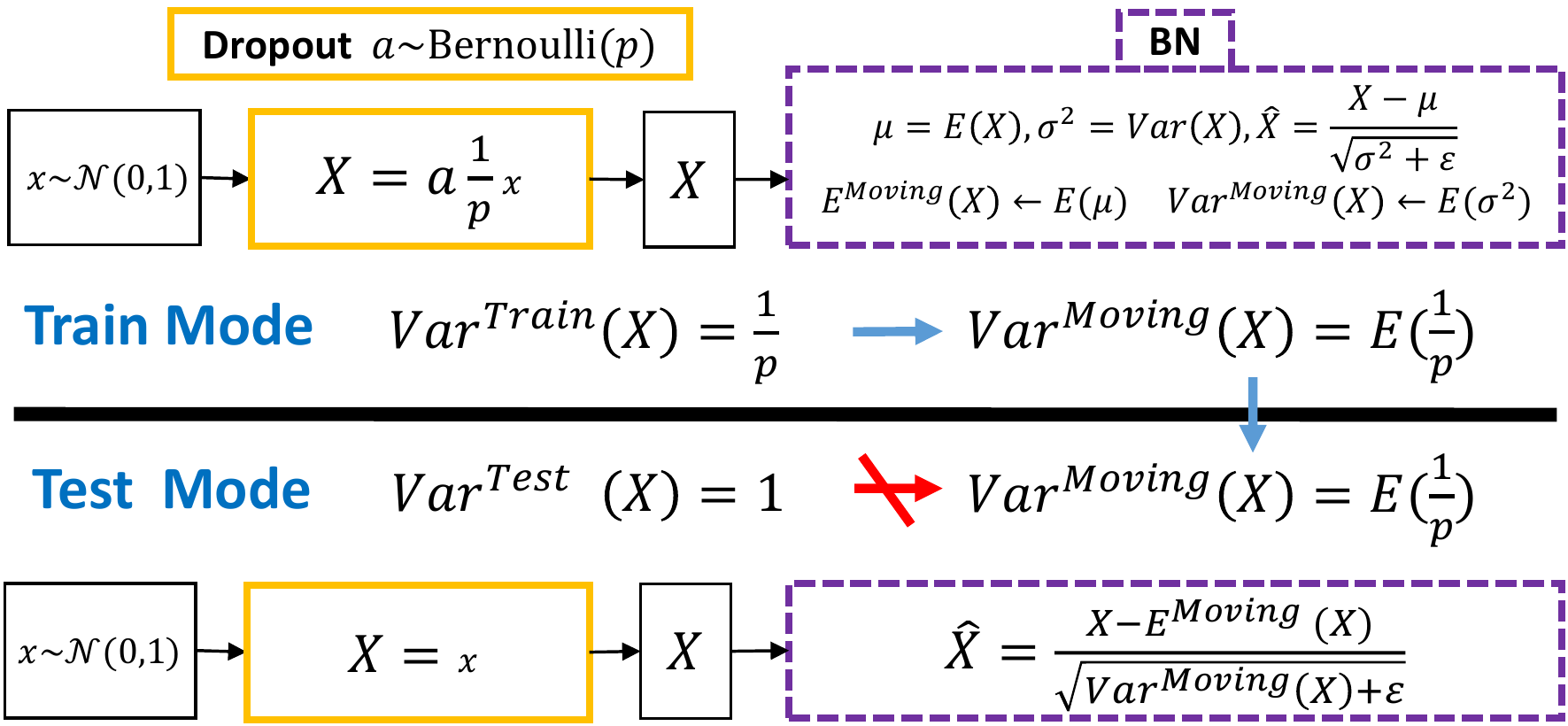}
	}

	\subfigure{
		\includegraphics[width=2.0in]{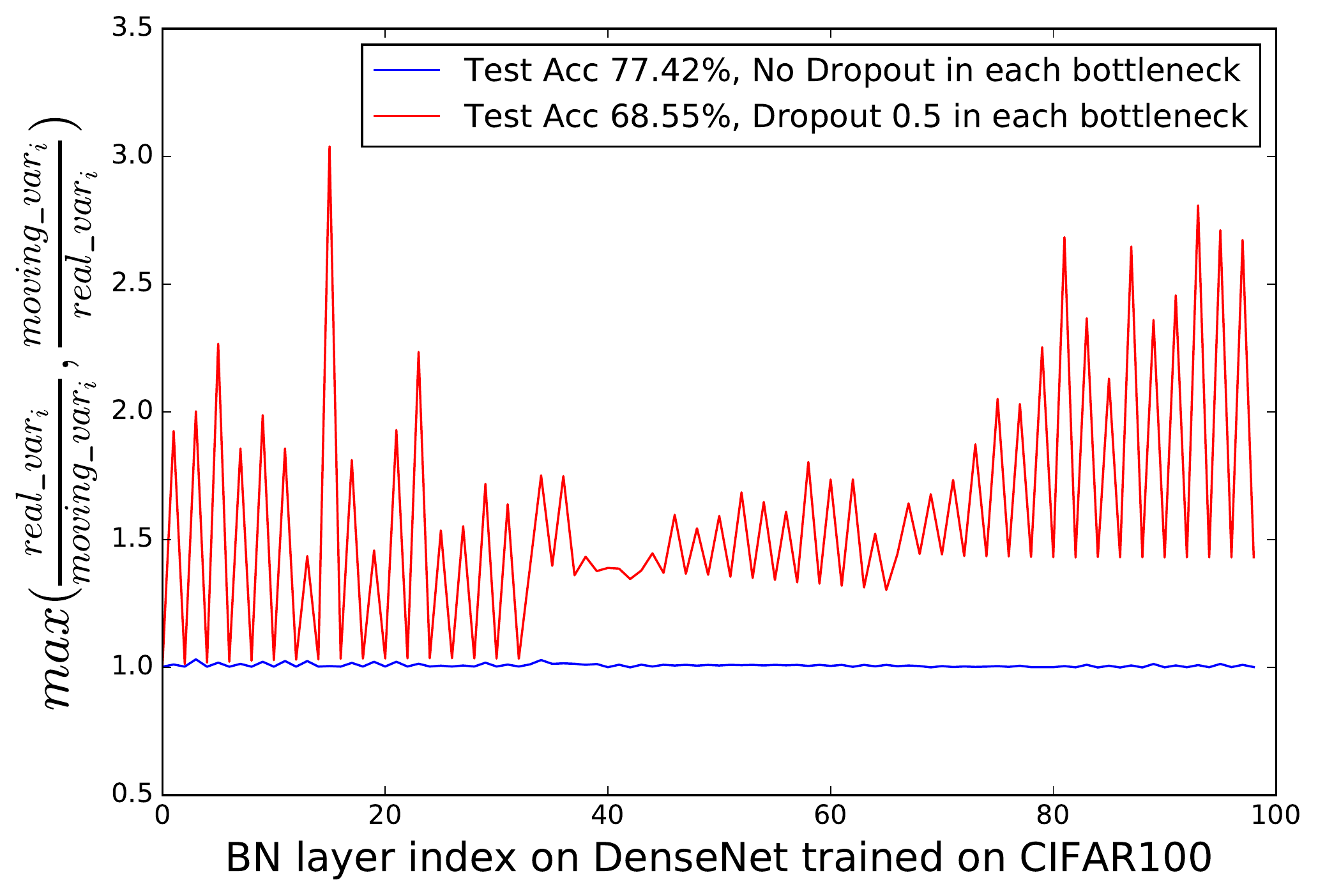}
	}
	\vspace{-10pt}
	\caption{\textbf{Up:} a simplified mathematical illustration of ``variance shift''. In test mode, the neural variance of $X$ is different from that in train mode caused by Dropout, yet BN attempts to regard that variance as the popular statistic accumulated from training. Note that $p$ denotes for the Dropout retain ratio and $a$ comes from Bernoulli distribution which has probability $p$ of being 1. \textbf{Down:} variance shift in experimental statistics on DenseNet trained on CIFAR100 dataset. The curves are both calculated from the \emph{same training data}. ``$\emph{moving\_var}_i$'' is the moving variance (take its mean value instead if it's a vector) that the $i$-th BN layer accumulates during the entire learning, and ``$\emph{real\_var}_i$'' stands for the real variance of neural response before the $i$-th BN layer in inference.}
	\vspace{-14pt}
	\label{bn_variance_shift}
\end{figure}

\vspace{-16pt}
\section{Introduction}

\cite{srivastava2014dropout} brought Dropout as a simple way to prevent neural networks from overfitting. It has been proved to be significantly effective over a large range of machine learning areas, such as image classification \cite{szegedy2015going}, speech recognition \cite{hannun2014deep} and even natural language processing \cite{kim2016character}. Before the birth of Batch Normalization, it became a necessity of almost all the state-of-the-art networks and successfully boosted their performances against overfitting risks, despite its amazing simplicity.

\cite{ioffe2015batch} demonstrated Batch Normalization (BN), a powerful skill that not only speeded up all the modern architectures but also improved upon their strong baselines by acting as regularizers. Therefore, BN has been implemented in nearly all the recent network structures \cite{szegedy2016rethinking,szegedy2017inception,Howard2017MobileNets,zhang2017shufflenet} and demonstrates its great practicability and effectiveness.


However, the above two nuclear weapons always fail to obtain an extra reward when combined together practically. In fact, a network even performs worse and unsatisfactorily when it is equipped with BN and Dropout simultaneously. \cite{ioffe2015batch} have already realized that BN eliminates the need for Dropout in some cases -- the authors exposed the incompatibility between them, thus conjectured that BN provides similar regularization benefits as Dropout intuitively. More evidences are provided in the modern architectures such as ResNet \cite{he2016deep,he2016identity}, ResNeXt \cite{xie2017aggregated}, DenseNet \cite{huang2016densely}, where the best performances are all obtained by BN with the absence of Dropout. Interestingly, a recent study Wide ResNet (WRN) \cite{zagoruyko2016wide} show that it is positive for Dropout to be applied in the WRN design with a large feature dimension. So far, previous clues leave us a mystery about the confusing and complicated relationship between Dropout and BN. Why do they conflict in most of the common architectures? Why do they cooperate friendly sometimes as in WRN? 

We discover the key to understand the disharmony between Dropout and BN is the inconsistent behaviors of neural variance during the switch of networks' state. Considering one neural response $X$ as illustrated in Figure~\ref{bn_variance_shift}, when the state changes from train to test, Dropout would scale the response by its Dropout retain ratio (i.e. $p$) that actually changes the neural variance as in learning, yet BN still maintains its statistical moving variance of $X$. This mismatch of variance could lead to a numerical instability (see \textcolor{red}{red} curve in Figure~\ref{bn_variance_shift}). As the signals go deeper, the numerical deviation on the final predictions may amplify, which drops the system's peformance. We name this scheme as ``variance shift'' for simplicity. Instead, without Dropout, the real neural variances in inference would appear very closely to the moving ones accumulated by BN (see \textcolor{blue}{blue} curve in Figure~\ref{bn_variance_shift}), which is also preserved with a higher test accuracy. 

Theoretically, we deduced the ``variance shift'' under two general conditions, and found a satisfied explanation for the aforementioned mystery between Dropout and BN. Further, a large range of experimental statistics from four modern networks (i.e., PreResNet \cite{he2016identity}, ResNeXt \cite{xie2017aggregated}, DenseNet \cite{huang2016densely}, Wide ResNet \cite{zagoruyko2016wide}) on the CIFAR10/100 datasets verified our findings as expected. 

Since the central reason for their performance drop was discovered, we adopted two strategies that explored the possibilities to overcome the limitation of their combination. One was to apply Dropout after all BN layers and another was to modify the formula of Dropout and made it less sensitive to variance. By avoiding the variance shift risks, most of them worked well and achieved extra improvements. 

\vspace{-3pt}
\section{Related Work and Preliminaries}
Dropout \cite{srivastava2014dropout} can be interpreted as a way of regularizing a neural network by adding noise to its hidden units. Specifically, it involves multiplying hidden activations by Bernoulli distributed random variables which take the value 1 with probability $p$ ($0 \le p \le 1$) and 0 otherwise\footnote{$p$ denotes for the Dropout retain ratio and $(1-p)$ denotes for the drop ratio in this paper.}. Importantly, the test scheme is quite different from the train. During training, the information flow goes through the dynamic sub-network. At test time, the neural responses are scaled by the Dropout retain ratio, in order to approximate an equally weighted geometric mean of the predictions of an exponential number of learned models that share parameters. Consider a feature vector $\mathbf{x} = (x_1\dots x_d)$ with channel dimension $d$. Note that this vector could be a part (one location) of convolutional feature-map or the output of the fully connected layer, i.e., it doesnot matter which type of network it lies in. If we apply Dropout on $\mathbf{x}$, for one unit ${x}_k, k = 1\dots d$, in the train phase, it is:
\begin{equation}
\widehat{x}_k = a_k x_k,
\end{equation} 
where $a_k \sim P$ that comes from the Bernoulli distribution:
\vspace{-4pt}
\begin{equation}
P(a_k)=\left\{
\begin{array}{ccl}
1-p, &  & {a_k = 0}\\
p, &  & {a_k = 1}\\	
\end{array} \right.
\label{eq_P},
\end{equation}
and $\mathbf{a} = (a_1\dots a_d)$ is a vector of \emph{independent} Bernoulli random variables. 
At test time for Dropout, one should scale down the weights by multiplying them by a factor of $p$. As introduced in \cite{srivastava2014dropout}, another way to achieve the same effect is to scale up the retained activations by multiplying by $\frac{1}{p}$ at training time and not modifying the weights at test time. It is more popular on practical implementations, thus we employ this formula of Dropout in both analyses and experiments. Therefore, the hidden activation in the train phase would be:
\vspace{-4pt}
\begin{equation}
\widehat{x}_k = a_k \frac{1}{p} x_k,
\end{equation} 
whilst in inference it would be simple like: $\widehat{{x}}_k = {x}_k$.

Batch Normalization (BN) \cite{ioffe2015batch} proposes a deterministic information flow by normalizing each neuron into zero mean and unit variance. Considering values of $x$ (for clarity, $x \equiv x_k$) over a mini-batch: $\mathcal{B} = \{x^{(1)...(m)}\}$\footnote{Note that we donot consider the ``scale and shift'' part in BN because the key of ``variance shift'' exists in its ``normalize'' part.} with $m$ instances, we have the form of ``normalize'' part:
\begin{equation}
\mu = \frac{1}{m}\sum_{i = 1}^{m}{x^{(i)}}, \sigma^2 = \frac{1}{m}\sum_{i = 1}^{m}{(x^{(i)} - \mu)^2}, \widehat{x}^{(i)} = \frac{x^{(i)} - \mu}{\sqrt{\sigma^2 + \epsilon}},  
\end{equation} 
where $\mu$ and $\sigma^2$ would participate in the backpropagation. The normalization of activations that depends on the mini-batch allows efficient training, but is neither necessary nor desirable during inference. Therefore, BN accumulates the moving averages of neural means and variances during learning to track the accuracy of a model as it trains:
\begin{equation}
E^{Moving}(x) \gets E_{\mathcal{B}}(\mu), {Var}^{Moving}(x) \gets E_{\mathcal{B}}^{'}(\sigma^2),
\end{equation} 
where $E_{\mathcal{B}}(\mu)$ denotes for the expectation of $\mu$ from multiple training mini-batches $\mathcal{B}$ and $E_{\mathcal{B}}^{'}(\sigma^2)$ denotes for the expectation of the unbiased variance estimate (i.e., $\frac{m}{m - 1}\cdot E_{\mathcal{B}}(\sigma^2)$) over multiple training mini-batches. They are all obtained by implementations of moving averages \cite{ioffe2015batch} and are fixed for linear transform during inference:
\begin{equation}
\widehat{x} = \frac{x - E^{Moving}(x)}{\sqrt{{Var}^{Moving}(x) + \epsilon}}.
\end{equation}  



\section{Theoretical Analyses}
\begin{figure}[ht]
	\begin{center}
		\centerline{\includegraphics[width=0.8\columnwidth]{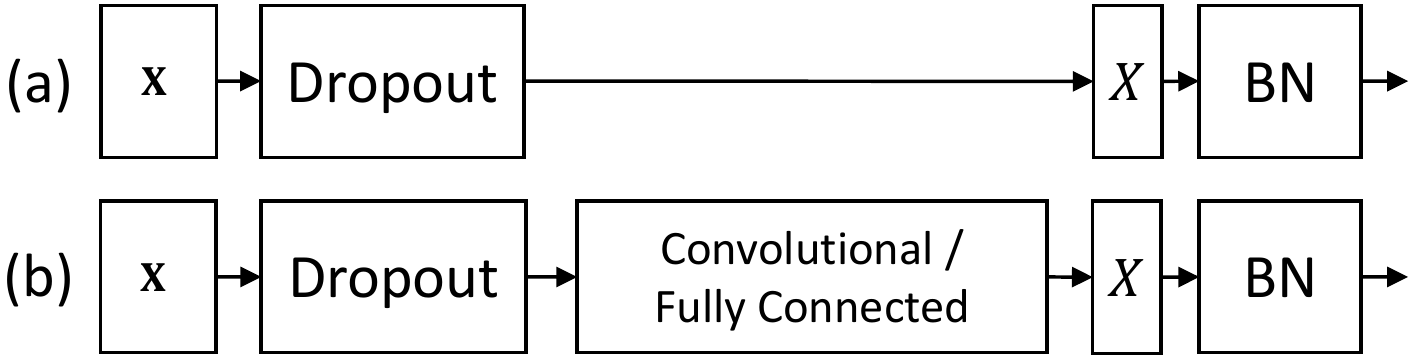}}
		\vskip -0.1in
		\caption{Two cases for analyzing variance shift.}
		\label{fig_twocases}
	\end{center}
	\vskip -0.2in
\end{figure}

From the preliminaries, one could notice that Dropout only ensures an ``equally weighted geometric \textbf{mean} of the predictions of an exponential number of learned models'' by the approximation from its test policy, as introduced in the original paper \cite{srivastava2014dropout}. This scheme poses the variance of the hidden units unexplored in a Dropout model. Therefore, the central idea is to investigate the variance of the neural response before a BN layer, where the Dropout is previously applied. This could be attributed into two cases generally, as shown in Figure \ref{fig_twocases}. 
In case (a), the BN layer is directly subsequent to the Dropout layer and we only need to consider one neural response $X = {a}_k \frac{1}{p} {x}_k, k = 1\dots d$ in train phase and $X = {x}_k$ in test phase. In case (b), the feature vector $\mathbf{x} = (x_1\dots x_d)$ would be passed into a convolutional layer (or a fully connected layer) to form the neural response $X$. We also regard its corresponding weights (the convolutional filter or the fully connected weight) to be $\mathbf{w} = (w_1\dots w_d)$, hence we get $X = \sum_{i = 1}^d{{w}_i {a}_i \frac{1}{p} {x}_i}$ for learning and $X = \sum_{i = 1}^d{{w}_i {x}_i}$ for test.
For the ease of deduction, we assume that the inputs all come from the distribution with $c$ mean and $v$ variance (i.e., $E(x_i) = c, Var(x_i)=v, i =1\dots d, v > 0$)
and we also start by studying the linear regime. We let the ${a}_i$ and ${x}_i$ be mutually independent, considering the property of Dropout. Due to the aforementioned definition, ${a}_i$ and ${a}_j$ are mutually independent as well.


\begin{figure*}[t]
	\begin{center}
		\centerline{\includegraphics[width=2.1\columnwidth]{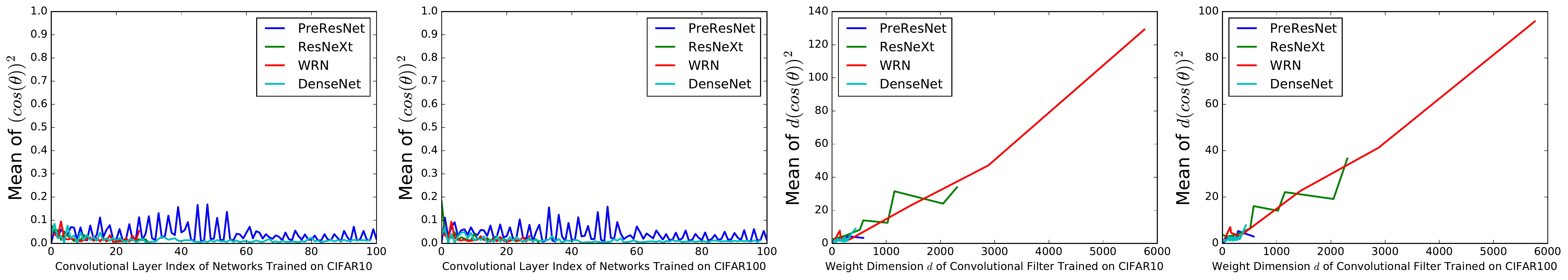}}
		\vskip -0.1in
		\caption{Statistical mean values of $({\cos \theta})^2$ and $d({\cos \theta})^2$. These four modern architectures are trained without Dropout on CIFAR10 and CIFAR100 respectively. We observe that $({\cos \theta})^2$ lies in $(0.01, 0.10)$ approximately in every network structure and various datasets. Interestingly, the term $d({\cos \theta})^2$ in WRN is significantly bigger than those on other networks mainly due to its larger channel width $d$.}
	\end{center}
	\vskip -0.2in
	\vspace{-10pt}
	\label{fig_mcos+cos_cropped}
\end{figure*}

\textbf{Figure \ref{fig_twocases} (a)}

Following the paradigms above, we have $Var^{Train}(X)$ as:
\begin{equation}
\begin{aligned}
&Var^{Train}(X) = Var(a_k \frac{1}{p} x_k) = E((a_k \frac{1}{p} x_k)^2) - E^2(a_k \frac{1}{p} x_k) \\
&= \frac{1}{p^2}E(a_k^2)E(x_k^2) - \frac{1}{p^2}(E(a_k)E(x_k))^2 = \frac{1}{p}(c^2+v) -c^2\\
\end{aligned}
\label{eq_a_var_train}
\end{equation}

In inference, BN keeps the moving average of variance (i.e., $E_{\mathcal{B}}^{'}(\frac{1}{p}(c^2+v) -c^2)$) fixed. In another word, BN wishes that the variance of neural response $X$, which comes from the input images, is supposed to be close to $E_{\mathcal{B}}^{'}(\frac{1}{p}(c^2+v) -c^2)$. However, Dropout breaks the harmony when it comes to its test stage by having $X = x_{k}$ to get $Var^{Test}(X)= Var(x_{k})=v$.
If putting $Var^{Test}(X)$ into the unbiased variance estimate, it would become $E_{\mathcal{B}}^{'}(v)$ which is obviously different from the popular statistic $E_{\mathcal{B}}^{'}(\frac{1}{p}(c^2+v) -c^2)$ of BN during training when Dropout ($p < 1$) is applied. Therefore, the shift ratio is obtained:
\vspace{-2pt}
\begin{equation}
\triangle(p)=\frac{Var^{Test}(X)}{Var^{Train}(X)} = \frac{v}{\frac{1}{p}(c^2+v) -c^2} 
\label{eq_a_var_train_test}
\end{equation}
In case (a), the variance shift happens via a coefficient $\triangle(p) \le 1$. Since modern neural networks carry a deep feedforward topologic structure, previous deviate numerical manipulations could lead to more uncontrollable numerical outputs of subsequent layers (Figure \ref{bn_variance_shift}). It brings the chain reaction of amplified shift of variances (even affects the means further) in every BN layers sequentially as the networks go deeper. We would show that it directly leads to a dislocation of final predictions and makes the system suffer from a performance drop later in the statistical experimental part (e.g., Figure \ref{fig_all}, \ref{fig_pics_wrong_cropped} in Section \ref{sec_exp}).

In this design (i.e., BN directly follows Dropout), if we want to alleviate the variance shift risks, i.e., $\triangle(p) \to 1$, the only thing we can do is to eliminate Dropout and set the Dropout retain ratio $p \to 1$. Fortunately, the architectures where Dropout brings benefits (e.g., in Wide ResNet) donot follow this type of arrangement. In fact, they adopt the case (b) in Figure \ref{fig_twocases}, which is more common in practice, and we would describe it in details next.

\textbf{Figure \ref{fig_twocases} (b)}

At this time, $X$ would be obtained by $\sum_{i = 1}^d{{w}_i {a}_i \frac{1}{p} {x}_i}$, where $\mathbf{w}$ denotes for the corresponding weights that act on the feature vector $\mathbf{x}$, along with the Dropout applied. For the ease of deduction, we assume that in the very later epoch of training, the weights of $\mathbf{w}$ remains constant given the gradients become significantly close to zero. Similarly, we can write $Var^{Train}(X)$ by following the formula of variance:
\begin{equation}
\begin{aligned}
&Var^{Train}(X) = Cov(\sum_{i = 1}^d{{w}_i {a}_i \frac{1}{p} {x}_i}, \sum_{i = 1}^d{{w}_i {a}_i \frac{1}{p} {x}_i}) \\
&= \frac{1}{p^2}\sum_{i=1}^{d}{({w}_i)^2Var({a}_i {x}_i)}\\
&+ \frac{1}{p^2}\sum_{i=1}^d\sum_{j \neq i}^{d}\rho_{i,j}^{ax}{w}_i{w}_j\sqrt{Var({a}_i {x}_i)}\sqrt{Var({a}_j {x}_j)} \\
&=(\frac{1}{p}(c^2+v) -c^2)(\sum_{i=1}^{d}{{w}_i^2}+\rho^{ax}\sum_{i=1}^d\sum_{j \neq i}^{d}{w}_i{w}_j),\\
\end{aligned}
\label{eq_b_var_train}
\end{equation}

where $\rho_{i,j}^{ax} = \frac{Cov({a}_i {x}_i,{a}_j {x}_j)}{\sqrt{Var({a}_i {x}_i)}\sqrt{Var({a}_j {x}_j)}} \in [-1, 1]$. For the ease of deduction, we simplify all the linear correlation coefficients to be the same as a constant $\rho^{ax} \cong \rho_{i,j}^{ax}, \forall i,j = 1\dots d, i \neq j$. Similarly, $Var^{Test}(X)$ is obtained:
\begin{equation}
\begin{aligned}
V&ar^{Test}(X) = Var(\sum_{i = 1}^{d}{ {w}_i {x}_i }) = Cov(\sum_{i = 1}^{d}{ {w}_i {x}_i },\sum_{i = 1}^{d}{ {w}_i {x}_i })\\
&=\sum_{i=1}^{d}{{w}_i^2}v + \sum_{i=1}^d\sum_{j \neq i}^{d}\rho_{i,j}^{x}{w}_i{w}_j\sqrt{v}\sqrt{v} \\
&=v(\sum_{i=1}^{d}{{w}_i^2} + \rho^{x}\sum_{i=1}^d\sum_{j \neq i}^{d}{w}_i{w}_j),
\end{aligned}
\label{eq_b_var_test}
\end{equation}
where $\rho_{i,j}^{x} = \frac{Cov({x}_i, {x}_j)}{\sqrt{Var( {x}_i)}\sqrt{Var( {x}_j)}} \in [-1, 1]$, and we also have a constant $\rho^{x} \cong \rho_{i,j}^{x}, \forall i,j = 1\dots d, i \neq j$. Since ${a}_i$ and ${x}_i$, ${a}_i$ and ${a}_j$ are mutually independent, we can get the relationship between $\rho^{ax}$ and $\rho^{x}$:
\begin{equation}
\begin{aligned}
&\rho^{ax} \cong \rho_{i,j}^{ax} = \frac{Cov({a}_i {x}_i,{a}_j {x}_j)}{\sqrt{Var({a}_i {x}_i)}\sqrt{Var({a}_j {x}_j)}}\\
&=\frac{p^2 Cov({x}_i, {x}_j)}{\frac{p(c^2+v)-p^2c^2}{v}\sqrt{Var({x}_i)}\sqrt{Var({x}_j)}}\\
&=\frac{v}{\frac{1}{p}(c^2+v) -c^2}\rho_{i,j}^{x} \cong \frac{v}{\frac{1}{p}(c^2+v) -c^2}\rho^{x}.
\end{aligned}
\label{eq_b_rho}
\end{equation}

According to Equation~\eqref{eq_b_var_train},~\eqref{eq_b_var_test} and \eqref{eq_b_rho}, we can write the variance shift $\frac{Var^{Test}(X)}{Var^{Train}(X)}$ as:
\begin{equation}
\begin{aligned}
&\frac{v(\sum_{i=1}^{d}{{w}_i^2} + \rho^{x}\sum_{i=1}^d\sum_{j \neq i}^{d}{w}_i{w}_j)}{(\frac{1}{p}(c^2+v) -c^2)(\sum_{i=1}^{d}{{w}_i^2}+\rho^{ax}\sum_{i=1}^d\sum_{j \neq i}^{d}{w}_i{w}_j)}\\
&=\frac{v\sum_{i=1}^{d}{{w}_i^2} + v\rho^x\sum_{i=1}^d\sum_{j \neq i}^{d}{w}_i{w}_j}{(\frac{1}{p}(c^2+v) -c^2)\sum_{i=1}^{d}{{w}_i^2} + v\rho^x\sum_{i=1}^d\sum_{j \neq i}^{d}{w}_i{w}_j}\\
&=\frac{v + v\rho^x((\sum_{i=1}^d{w_i})^2 - \sum_{i=1}^{d}{{w}_i^2})/\sum_{i=1}^{d}{{w}_i^2}}{\frac{1}{p}(c^2+v) -c^2 + v\rho^x((\sum_{i=1}^d{w_i})^2 - \sum_{i=1}^{d}{{w}_i^2})/\sum_{i=1}^{d}{{w}_i^2}}\\
&=\frac{v + v\rho^x(d(\cos\theta)^2 - 1)}{\frac{1}{p}(c^2+v) -c^2 + v\rho^x(d(\cos\theta)^2 - 1)},
\end{aligned}
\label{eq_b_div}
\end{equation}
\begin{table}[t]
	\vspace{-10pt}
	\caption{Statistical means of $({\cos \theta})^2$ and $d({\cos \theta})^2$ over all the convolutional layers on four representative networks.}
	\centering
	\begin{small}
		\begin{tabular}{lcrcr}
			\toprule
			\multirow{2}{*}{Networks} & \multicolumn{2}{c}{CIFAR10} & \multicolumn{2}{c}{CIFAR100} \\
			\cline{2-5}
			\specialrule{0em}{1pt}{1pt}
			&$({\cos \theta})^2$ & $d({\cos \theta})^2$  &$({\cos \theta})^2$ & $d({\cos \theta})^2$\\
			\midrule
			PreResNet&  0.03546& 2.91827 & 0.03169 & 2.59925\\
			ResNeXt  &  0.02244& 14.78266& 0.02468 & 14.72835\\
			WRN      &  0.02292& 52.73550& 0.02118 & 44.31261\\
			DenseNet &  0.01538& 3.83390 & 0.01390 & 3.43325\\
			\bottomrule	
		\end{tabular}
	\end{small}
	\vspace{-30pt}
	\label{tab_summary}
\end{table}

where $({\cos \theta})^2$ comes from the expression:
\begin{equation}
\begin{aligned}
\frac{(\sum_{i=1}^{d}{{w}_i})^2}{d \cdot\sum_{i=1}^{d}{{w}_i^2}} 
&= {( \frac{\sum_{i=1}^{d}{1 \cdot {w}_i}}{\sqrt{\sum_{i=1}^{d}{1^2}}\sqrt{\sum_{i=1}^{d}{{w}_i^2}}})}^2 = ({\cos \theta})^2,\\
\end{aligned}
\label{eq_cos}
\end{equation}
and $\theta$ denotes for the angle between vector $\mathbf{w}$ and vector $\underbrace{(1\dots 1)}_{\text{m}}$. To prove that $d({\cos \theta})^2$ scales approximately linear to $d$, we made rich calculations w.r.t the term $d({\cos \theta})^2$ and $({\cos \theta})^2$ on four modern architectures\footnote{For the convolutional filters which have larger than $1$ filter size as $k \times k, k > 1$, we vectorise them by expanding its channel width $d$ to $d \times k \times k$ while maintaining all the weights.} trained on CIFAR10/100 datasets (Table \ref{tab_summary} and Figure \ref{fig_mcos+cos_cropped}). Based on Table \ref{tab_summary} and Figure \ref{fig_mcos+cos_cropped}, we observe that $({\cos \theta})^2$ lies in $(0.01, 0.10)$ stably in every network and various datasets whilst $d({\cos \theta})^2$ tends to increase in parallel when $d$ grows. From Equation~\eqref{eq_b_div}, the inequation $Var^{Test}(X) \neq Var^{Train}(X)$ holds when $p < 1$. If we want $Var^{Test}(X)$ to approach $Var^{Train}(X)$, we need this term 
\begin{equation}
\begin{aligned}
\triangle(p, d)&=\frac{Var^{Test}(X)}{Var^{Train}(X)}= \frac{v\rho^x(d(\cos\theta)^2 - 1)+ v}{ v\rho^x(d(\cos\theta)^2 - 1) + \frac{1}{p}(c^2+v) -c^2}\\
&= \frac{v\rho^x + \frac{v(1 - \rho^x)}{d(\cos\theta)^2}}{v\rho^x + \frac{(\frac{1}{p} - 1)c^2 + v(\frac{1}{p} - \rho^x)}{d(\cos\theta)^2}}
\end{aligned}
\end{equation}
to approach $1$. There are \emph{two ways} to achieve $\triangle(p, d) \to 1$: 
\begin{itemize}
	\item $p \to 1$: maximizing the Dropout retain ratio $p$ (ideally up to $1$ which means Dropout is totally eliminated); 
	\item $d \to \infty$: growing the width of channel exactly as the Wide ResNet did to enlarge $d$.
\end{itemize}

\section{Statistical Experiments}
\label{sec_exp}

We conduct extensive statistical experiments to check the correctness of above deduction in this section. Four modern architectures including DenseNet \cite{huang2016densely}, PreResNet \cite{he2016identity}, ResNeXt \cite{xie2017aggregated} and Wide ResNet (WRN) \cite{zagoruyko2016wide} are adopted on the CIFAR10 and CIFAR100 datasets. 

\textbf{Datasets.} The two CIFAR datasets \cite{krizhevsky2009learning} consist of colored natural scence images, with 32$\times$32 pixel each. The train and test sets contain 50k images and 10k images respectively. CIFAR10 (C10) has 10 classes and CIFAR100 (C100) has 100. For data preprocessing, we normalize the data by using the channel means and standard deviations. For data augmentation, we adopt a standard scheme that is widely used in \cite{he2016identity,huang2016densely,larsson2016fractalnet,lin2013network,lee2015deeply,springenberg2014striving,srivastava2015training}: the images are first zero-padded with 4 pixels on each side, then a 32$\times$32 crop is randomly sampled from the padded images and at least half of the images are horizontally flipped. 

\textbf{Networks with Dropout.} The four modern architectures are all chosen from the open-source codes\footnote{Our implementations basicly follow the public code in https://github.com/bearpaw/pytorch-classification. The training details can also be found there. Our code for the following experiments would be released soon.} written in pytorch that can reproduce the results reported in previous papers. The details of the networks are listed in Table \ref{tab_network_details}:
\begin{table}[h]
	\vspace{-10pt}
	\caption{Details of four modern networks in experiments. \#P denotes for the amount of model parameters.}
	\centering
	\begin{small}
		\begin{tabular}{lrr}
			\toprule
			Model  & \#P on C10 & \#P on C100 \\
			\midrule
			PreResNet-110				& 1.70 M 	& 1.77 M \\
			ResNeXt-29, 8 $\times$ 64  	& 34.43 M  	& 34.52 M  \\
			WRN-28-10      				& 36.48 M	& 36.54 M \\
			DenseNet-BC (L=100, k=12) 	& 0.77 M 	& 0.80 M \\
			\bottomrule	
		\end{tabular}
	\end{small}
	\label{tab_network_details}
	\vspace{-10pt}
\end{table}

Since the BN layers are already developed as the indispensible components of their body structures, we arrange Dropout that follows the two cases in Figure \ref{fig_twocases}:

\textbf{(a)} We assign all the Dropout layers only and right before all the bottlenecks' last BN layers in these four networks, neglecting their possible Dropout implementations (as in DenseNet \cite{huang2016densely} and Wide ResNet \cite{zagoruyko2016wide}). We denote this design to be models of \textbf{Dropout-(a)}.

\textbf{(b)} We follow the assignment of Dropout in Wide ResNet \cite{zagoruyko2016wide}, which finally improves WRNs' overall performances, to place the Dropout before the last Convolutional layer in every bottleneck block of PreResNet, ResNeXt and DenseNet. This scheme is denoted as \textbf{Dropout-(b)} models.


\textbf{Statistics of variance shift.} Assume a network $\mathcal{G}$ contains $n$ BN layers in total. We arrange these BN layers from shallow to deep by giving them indices that goes from $1$ to $n$ accordingly. The whole statistical manipulation is conducted by following three steps:

\textbf{(1) Calculate \emph{moving\_var}$_i, i \in [1, n]$}: when $\mathcal{G}$ is trained until convergence, each BN layer obtains the moving average of neural variance (the unbiased variance estimate) from the feature-map that it receives during the entire learning procedure. We denote that variance to be \emph{moving\_var}. Since the \emph{moving\_var} for every BN layer is a vector (whose length is equal to the amount of channels of previous feature-map), we leverage its mean value to represent \emph{moving\_var} instead, for a better visualization. Further, we denote \emph{moving\_var}$_i$ as the \emph{moving\_var} of $i$-th BN layer.


\textbf{(2) Calculate \emph{real\_var}$_i, i \in [1, n]$}: after training, we fix all the parameters of $\mathcal{G}$ and set its state to be the evaluation mode (hence the Dropout would apply its inference policy and BN would freeze its moving averages of means and variances). The training data is again utilized for going through $\mathcal{G}$ within a certain of epochs, in order to get the real expectation of neural variances on the feature-maps before each BN layer. Data augmentation is also kept to ensure that every possible detail for calculating neural variances remains exactly the same with training. Importantly, we adopt the same moving average algorithm to accumulate the unbiased variance estimates. Similarly in (1), we let the mean value of real variance vector be \emph{real\_var}$_i$ before the $i$-th BN layer. 

\begin{figure*}[t]
	\begin{center}
		\centerline{\includegraphics[width=2.1\columnwidth]{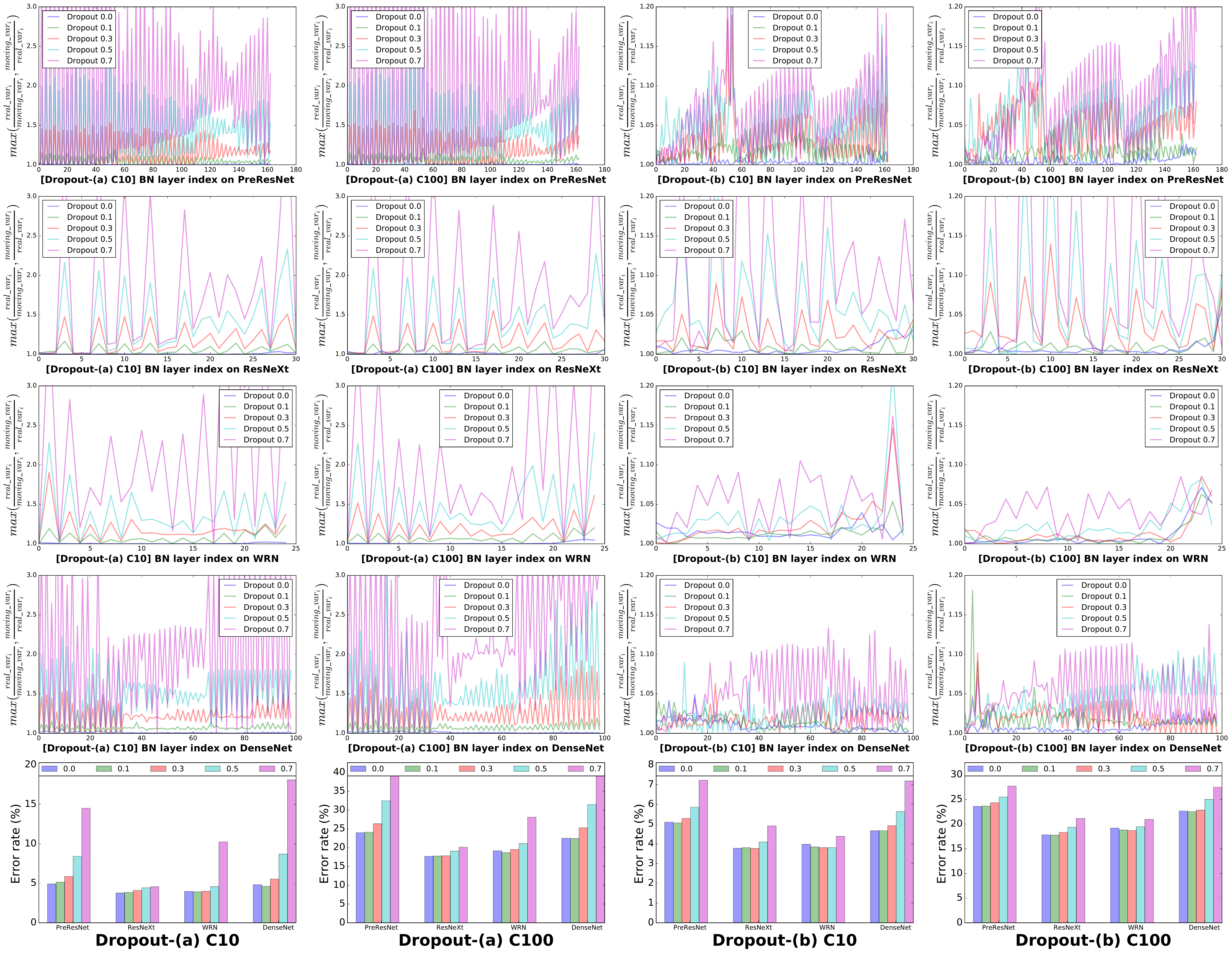}}
		\vskip -0.1in
		\caption{\textbf{See by columns.} Statistical visualizations about ``variance shift'' on BN layers of four modern networks w.r.t: 1) Dropout type; 2) Dropout drop ratio; 3) dataset, along with their test error rates (the fifth row). Obviously, WRN is less influenced by Dropout (i.e., small variance shift) when the Dropout-(b) drop ratio $\le 0.5$, thus it even enjoys an improvement with Dropout applied before BN.}
		\label{fig_all}
	\end{center}
	\vspace{-25pt}
\end{figure*}

\textbf{(3) Obtain $\emph{max}(\frac{\emph{real\_var}_i}{\emph{moving\_var}_i},\frac{\emph{moving\_var}_i}{\emph{real\_var}_i}), i \in [1, n]$}: since we focus on the shift, the scalings are all kept above $1$ by their reciprocals if possible in purpose of a better view. Various Dropout drop ratios $[0.0, 0.1, 0.3, 0.5, 0.7]$ are applied for clearer comparisons in Figure \ref{fig_all}. The corresponding error rates are also included in each column. 

\textbf{Agreements between analyses and experiments about the relation between performance and variance shift.} In these four columns of Figure \ref{fig_all}, we discover that when the drop ratio is relatively small (i.e., 0.1), the \textcolor{green}{green} shift curves are all near the \textcolor{blue}{blue} ones (i.e. models without Dropout), thus their performances are as well very close to the baselines. It agrees with our previous deduction that whenever in (a) or (b) case, decreasing drop ratio $1-p$ would alleviate the variance shift risks. Furthermore, in Dropout-(b) models (i.e., the last two columns) we find that, for WRNs, the curves with drop ratio $0.1, 0.3$ even $0.5$ approaches closer to the one with $0.0$ than other networks, and they all outperform the baselines. It also aligns with our analyses since WRN has a significantly larger channel dimension $d$, and it ensures that a slightly larger $p$ would not explode the neural variance but bring the original benefits, which Dropout carries, back to the BN-equipped networks.  

\begin{figure*}[t]
	\begin{center}
		\centerline{\includegraphics[width=2\columnwidth]{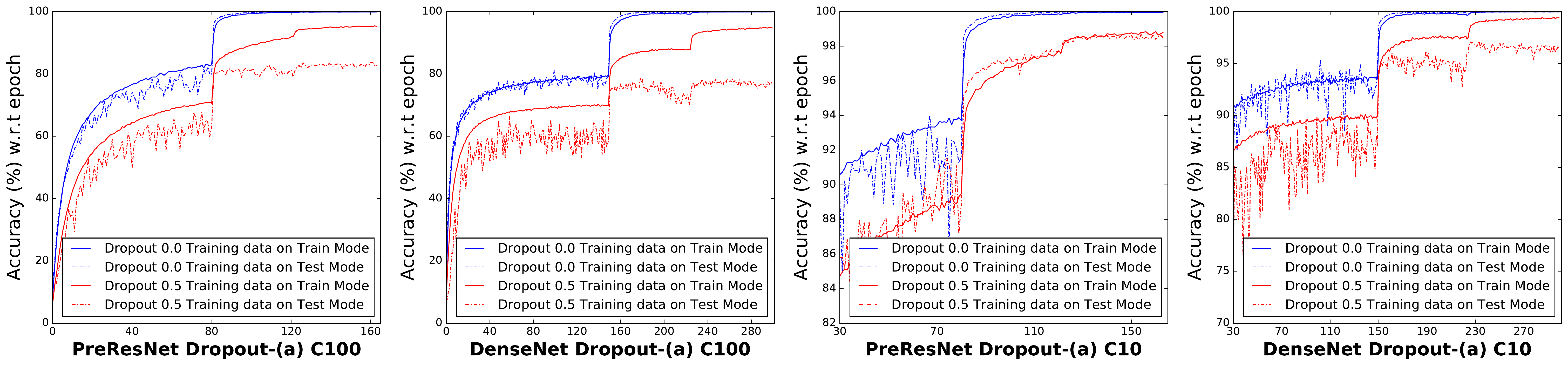}}
		\vskip -0.1in
		\caption{Accuracy by train epochs. Curves in \textcolor{blue}{blue} means the train of these two networks without Dropout. Curves in \textcolor{red}{red} denotes the Dropout version of the corresponding models. These accuracies are all calculated from the training data, while the solid curve is under train mode and the dashed one is under evaluation mode. We observe the significant accuracy shift when a network with Dropout ratio $0.5$ changes its state from train to test stage, with all network parameters fixed but the test policies of Dropout and BN applied. }
		\label{fig_train_train_cropped}
	\end{center}
	\vspace{-17pt}
\end{figure*}
\begin{figure*}[t]
	\begin{center}
		\centerline{\includegraphics[width=2\columnwidth]{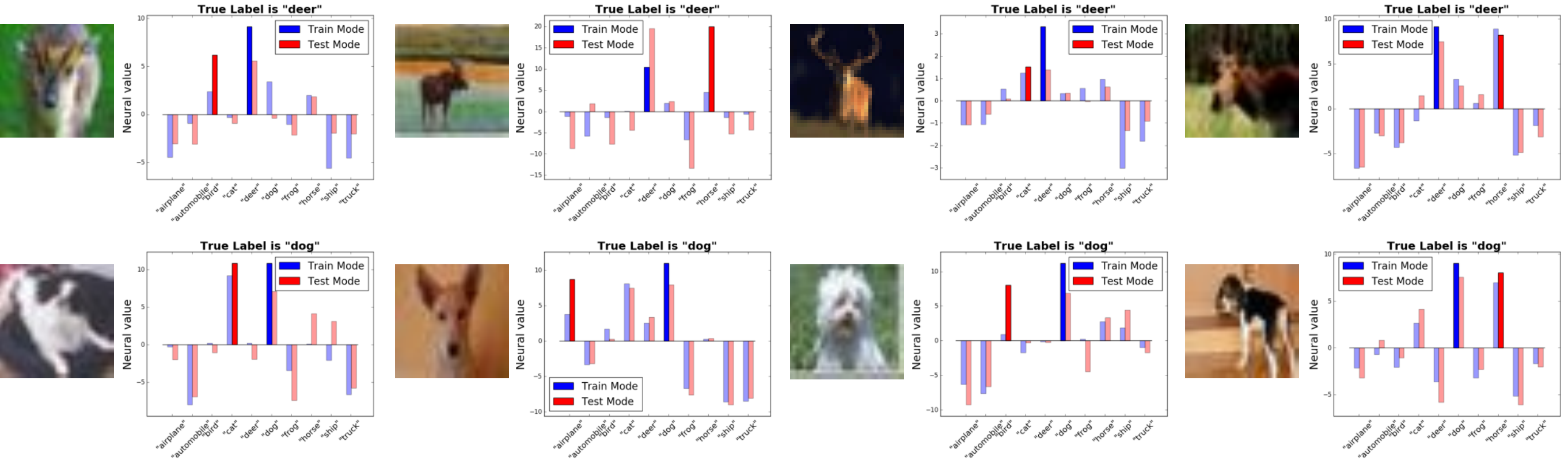}}
		\vskip -0.1in
		\caption{Examples of inconsistent neural responses between train mode and test mode of DenseNet Dropout-(a) $0.5$ trained on CIFAR10 dataset. These samples are from the training data, whilst they are correctly classified by the model during learning yet erroneously judged in inference, despite all the fixed model parameters. Variance shift finally leads to the prediction shift that drops the performance.}
		\label{fig_pics_wrong_cropped}
	\end{center}
	\vskip -0.2in
	\vspace{-13pt}
\end{figure*}

\textbf{Even the training data performs inconsistently between train and test mode.} In addition, we also observe that for DenseNet and PreResNet (their channel $d$ is relatively small), when their state is changed from train to test, even the training data cannot be kept with a coherent accuracy at last. In inference, the variance shift happens and it leads to an avalanche effect on the numerical explosion and instability in networks that finally changes the final prediction. Here we take the two models with drop ratio being $0.5$ as an example, hence demonstrate that a large amount of training data would be classified inconsistently between train and test mode, despite their same model parameters (Figure \ref{fig_train_train_cropped}). 

\textbf{Neural responses (of last layer before softmax) for training data are unstable from train to test.} To get a clearer understanding of the numerical disturbance that the variance shift brings finally, a bundle of images (from training data) are drawn with their neural responses before the softmax layer in both train stage and test stage (Figure \ref{fig_pics_wrong_cropped}). From those pictures and their responses, we can find that with all the weights of networks fixed, only a mode transfer (from train to test) would change the distribution of the final responses even in the train set, and it leads to a wrong classification consequently. It proves that the predictions of training data differs between train stage and test stage when a network is equipped with Dropout layers before BN layers. Therefore, we confirm that the unstable numerical behaviors are the fundamental reasons for the performance drop.
\begin{table}[t]
	\vspace{-8pt}
	\caption{Adjust BN's moving mean/variance by running moving average algorithm on training data under test mode. These numbers are all averaged from $5$ parallel runnings with different random initial seeds. \textcolor{red}{\textbf{}}}
	\centering
	\begin{small}
		\begin{tabular}{lcc|cc}
			\toprule
			\multirow{2}{*}{\ \ \ \ C10} & \multicolumn{2}{c}{Dropout-(a)} & \multicolumn{2}{c}{Dropout-(b)}\\
			\cline{2-5}
			\specialrule{0em}{1pt}{1pt}
			&0.5&0.5-Adjust&0.5&0.5-Adjust\\
			\midrule
			PreResNet & 8.42& \textbf{6.42}& 5.85&\textbf{5.77}\\
			ResNeXt   & 4.43& \textbf{3.96}& 4.09&\textbf{3.93}\\
			WRN       & 4.59& \textbf{4.20}& 3.81&\textbf{3.71}\\
			DenseNet  & 8.70& \textbf{6.82}& 5.63&\textbf{5.29}\\
			\midrule
			\multirow{2}{*}{\ \ \ \ C100} & \multicolumn{2}{c}{Dropout-(a) } & \multicolumn{2}{c}{Dropout-(b)}\\
			\cline{2-5}
			\specialrule{0em}{1pt}{1pt}
			&0.5&0.5-Adjust&0.5&0.5-Adjust\\
			\midrule
			PreResNet & 32.45& \textbf{26.57}& 25.50& \textbf{25.20}\\
			ResNeXt   & 19.04& \textbf{18.24}& 19.33&
			\textbf{19.09}\\
			WRN       & 21.08& \textbf{20.70}& 19.48& \textbf{19.15}\\
			DenseNet  & 31.45& \textbf{26.98}& 25.00& \textbf{23.92}\\
			\bottomrule	
		\end{tabular}
	\end{small}
	\vspace{-16pt}
	\label{tab_change_BN_1}
\end{table}

\textbf{Only an adjustment for moving means and variances would bring an improvement, despite all other parameters fixed.} Given that the moving means and variances of BN would not match the real ones during test, we attempt to adjust these values by passing the training data again under the evaluation mode. In this way, the moving average algorithm \cite{ioffe2015batch} can also be applied. After shifting the moving statistics to the real ones by using the training data, we can have the model performed on the test set. From Table \ref{tab_change_BN_1}, All the Dropout-(a)/(b) $0.5$ models outperform their baselines by having their moving statistics adjusted. Significant improvements (e.g., $\sim 2$ and $\sim 4.5$ gains for DenseNet on CIFAR10 and on CIFAR100 respectively) can be observed in Dropout-(a) models. It again verifies that the drop of performance could be attributed to the ``variance shift'': a more proper popular statistics with \emph{smaller} variance shift could recall a bundle of erroneously classified samples back to right ones.


%


\section{Strategies to Combine Them Together}
Since we get a clear knowledge about the disharmony between Dropout and BN, we can easily develop several approaches to combine them together, to see whether an extra improvement could be obtained. In this section, we introduce two possible solutions in modifying Dropout. One is to avoid the scaling on feature-map before every BN layer, by only applying Dropout after the last BN block. Another is to slightly modify the formula of Dropout and make it less sensitive to variance, which can alleviate the shift problem and stabilize the numerical behaviors.   

\begin{table}[t]
	\vspace{-8pt}
	\centering
	\caption{Error rates after applying Dropout after all BN layers. These numbers are all averaged from $5$ parallel runnings with different random initial seeds. \textcolor{red}{\textbf{}}}
	\begin{small}
		\begin{tabular}{lccccc}
			\toprule
			C10 \ drop ratio&0.0&0.1&0.2&0.3&0.5\\
			\midrule
			PreResNet & 5.02 &4.96&5.01&\textbf{4.94}&5.03\\
			ResNeXt   & 3.77 &3.89&\textbf{3.69}&3.78&3.78\\
			WRN       & 3.97 &3.90&4.00&3.93&\textbf{3.84}\\
			DenseNet  & 4.72 &\textbf{4.67}&4.73&4.75&4.87 \\
			\midrule
			C100 \ drop ratio&0.0&0.1&0.2&0.3&0.5\\
			\midrule
			PreResNet & 23.73 &\textbf{23.43}&23.65&23.45&23.76\\
			ResNeXt   & 17.78 &\textbf{17.77}&17.99&17.97&18.26\\
			WRN       & \textbf{19.17} &\textbf{19.17}&19.23&19.19&19.25\\
			DenseNet  & 22.58 &\textbf{21.86}&22.41&22.41&23.49\\
			\bottomrule	
		\end{tabular}
	\end{small}
	\label{tab_change_dropout_1}
	\vspace{-13pt}
\end{table}

\textbf{Apply Dropout after all BN layers.} According to above analyses, the variance shift only happens when there exists a Dropout layer before a BN layer. Therefore, the most direct and concise way to tackle this is to assign Dropout in the position where the subsequent layers donot include BN. Inspired by early works that applied Dropout on the fully connected layers in \cite{krizhevsky2012imagenet}, we add only one Dropout layer right before the \emph{softmax} layer in these four architectures. Table \ref{tab_change_dropout_1} indicates that such a simple operation could bring $0.1$ improvements on CIFAR10 and reach up to $0.7$ gain on CIFAR100 for DenseNet. Please note that the last-layer Dropout performs worse on CIFAR100 than on CIFAR10 generally since the training data of CIFAR100 is insufficient and these models may suffer from certain underfitting risks. We also find it interesting that WRN may not need to apply Dropout on each bottleneck block -- only a last Dropout layer could bring enough or at least comparable benefits on CIFAR10. Additionally, we discover that in some previous work like \cite{hu2017squeeze}, the authors already adopted the same tips in their winning solution on the ILSVRC 2017 Classification Competition. Since it didnot report the gain that last-layer Dropout brings, we made some additional experiments and evaluate several state-of-the-art models on the ImageNet \cite{russakovsky2015imagenet} validation set (Table \ref{tab_change_dropout_1_imagenet}) using a 224 $\times$ 224 centre crop evaluation on each image (where the shorter edge is first resized to 256). We observe consistent improvements when drop ratio $0.2$ is employed after all BN layers on the large scale dataset.
\begin{table}[t]
	\vspace{-8pt}
	\centering
	\caption{Error rates after applying Dropout after all BN layers on the representative state-of-the-art models on ImageNet. These numbers are averaged from $5$ parallel runnings with different random initial seeds.  Consistent improvements can be observed.\textcolor{red}{\textbf{}}}
	\begin{small}
		\begin{tabular}{lcccc}
			\toprule
			\multirow{2}{*}{\shortstack{ \\\\
					ImageNet \ \ drop ratio}} & \multicolumn{2}{c}{top-1} &\multicolumn{2}{c}{top-5} \\
			\cline{2-5}
			\specialrule{0em}{1pt}{1pt}
			&0.0&0.2&0.0&0.2\\
			\midrule
			ResNet-200 \cite{he2016identity}    & 21.70& \textbf{21.48} &5.80 & \textbf{5.55}\\
			ResNeXt-101\cite{xie2017aggregated} & 20.40 & \textbf{20.17} & 5.30  & \textbf{5.12}\\
			SENet \cite{hu2017squeeze} 			& 18.89 & \textbf{18.68} & 4.66 & \textbf{4.47}\\
			\bottomrule	
		\end{tabular}
	\end{small}
	\label{tab_change_dropout_1_imagenet}
	\vspace{-15pt}
\end{table}

\textbf{Change Dropout into a more variance-stable form.} The drawbacks of vanilla Dropout lie in the weight scale during the test phase, which may lead to a large disturbance on statistical variance. This clue could push us to think: if we find a scheme that functions like Dropout but carries a lighter variance shift, we may stabilize the numerical behaviors of neural networks, thus the final performance would probably enjoy a possible benefit. Here we take the \textbf{Figure \ref{fig_twocases} (a)} case as an example for investigation where the variance shift rate is $\frac{v}{\frac{1}{p}(c^2+v) -c^2} = p$ (we let $c = 0$ for simplicity). That is, if we set the drop ratio $(1-p)$ as $0.1$, the variance would be scaled by $0.9$ when the network is transferred from train to test. Inspired by the original Dropout paper \cite{srivastava2014dropout} where the authors also proposed another form of Dropout that amounts to adding a Gaussian distributed random variable with zero mean and standard deviation equal to the activation of the unit, i.e., ${x}_i + {x}_ir$ and $r \sim \mathcal{N}(0, 1)$, we modify $r$ as a uniform distribution that lies in $[-\beta, \beta]$, where $0 \le \beta \le 1$. Therefore, each hidden activation would be $X = {x}_i + {x}_i{r}_i$ and ${r}_i \sim \mathcal{U}(-\beta, \beta)$. We name this form of Dropout as ``Uout'' for simplicity. With the mutually independent distribution between ${x}_i$ and ${r}_i$ being hold, we apply the form $X = {x}_i + {x}_i{r}_i$, ${r}_i \sim \mathcal{U}(-\beta, \beta)$ in train stage and $X = {x}_i$ in test mode. Similarly, in the simplified case of $c = 0$, we can deduce the variance shift again as follows:
\vspace{-6pt}
\begin{equation}
\begin{aligned}
&\frac{Var^{Test}(X)}{Var^{Train}(X)} = \frac{Var({x}_i)}{Var({x}_i + {x}_i{r}_i)} = \frac{v}{E(({x}_i + {x}_i{r}_i)^2)} \\
&=\frac{v}{E({x}_i^2) + 2E({x}_i^2)E({r}_i) + E({x}_i^2)E({r}_i^2)} = \frac{3}{3 + \beta^2}.
\end{aligned}
\end{equation}
Giving $\beta$ as $0.1$, the new variance shift rate would be $\frac{300}{301} \approx 0.9966777$ which is much closer to $1.0$ than the previous $0.9$ in \textbf{Figure \ref{fig_twocases} (a)}. A list of experiments is hence employed based on those four modern networks under {Dropout-(b)} settings w.r.t $\beta$ (Table \ref{tab_change_dropout_2}). We find that ``Uout'' would be less affected by the insufficient training data on CIFAR100 than applying the last-layer Dropout, which indicates a superior property of stability. Except for ResNeXt, nearly all the architectures achieved up to $0.2\sim 0.3$ increase of accuracy on both CIFAR10 and CIFAR100 dataset. 
\begin{table}[t]
	\vspace{-8pt}
	\caption{Apply new form of Dropout (i.e. Uout) in {Dropout-(b)} models. These numbers are all averaged from $5$ parallel runnings with different random initial seeds. \textcolor{red}{\textbf{}}}
	\centering
	\begin{small}
		\begin{tabular}{lccccc}
			\toprule
			C10 \ \ $\beta$&0.0&0.1&0.2&0.3&0.5\\
			\midrule
			PreResNet & 5.02 &5.02&\textbf{4.85}&4.98&4.97\\
			ResNeXt   & 3.77 &3.84&3.83&\textbf{3.75}&3.79\\
			WRN       & 3.97 &3.96&\textbf{3.80}&3.90&3.84\\
			DenseNet  & 4.72 &4.70&4.64&4.68&\textbf{4.61} \\
			\midrule
			C100 \ \ $\beta$&0.0&0.1&0.2&0.3&0.5\\
			\midrule
			PreResNet & 23.73 &23.73&23.62&\textbf{23.53}&23.77\\
			ResNeXt   & 17.78 &\textbf{17.74}&17.77&17.83&17.86\\
			WRN       & 19.17 &19.07&18.98&18.95&\textbf{18.87}\\
			DenseNet  & 22.58 &22.39&22.57&22.35&\textbf{22.30}\\
			\bottomrule	
		\end{tabular}
	\end{small}
	\vspace{-14pt}
	\label{tab_change_dropout_2}
\end{table}

\section{Conclusion}
In this paper, we investigate the ``variance shift'' phenomenon when Dropout layers are applied before Batch Normalization on modern neural networks. We discover that due to their distinct test policies, neural variance would be improper and shifted as the information flows in inference, and it leads to the unexpected final predictions that drops the performance. To avoid the variance shift risks, we next explore two strategies, and they are proved to work well in practice. We highly recommand that researchers could take these solutions to boost their models' performance if further improvement is desired, since their extra cost is nearly free and they are easy to be implemented.

\bibliography{example_paper}
\bibliographystyle{icml2018}

\end{document}